\documentclass{article}

\PassOptionsToPackage{numbers, compress}{natbib}
\usepackage[preprint]{neurips_2023}

\usepackage[utf8]{inputenc} % allow utf-8 input
\usepackage[T1]{fontenc}    % use 8-bit T1 fonts
\usepackage{hyperref}       % hyperlinks
\usepackage{url}            % simple URL typesetting
\usepackage{booktabs}       % professional-quality tables
\usepackage{amsfonts}       % blackboard math symbols
\usepackage{nicefrac}       % compact symbols for 1/2, etc.
\usepackage{microtype}      % microtypography
\usepackage{xcolor}         % colors
\usepackage{graphicx}
\usepackage{natbib}

\title{LLM-BRAIn: AI-driven Fast Generation of Robot Behaviour Tree based on Large Language Model}

\author{
  Artem Lykov\\
  Department of Engineering Systems\\
  Skolkovo Institute of Science and Technology\\
  Moscow Oblast, 121205, Russia \\
  \texttt{artem.lykov@skoltech.ru} \\
  \And Dzmitry Tsetserukou\\
  Department of Engineering Systems\\
  Skolkovo Institute of Science and Technology\\
  Moscow Oblast, 121205, Russia \\
  \texttt{d.tsetserukou@skoltech.ru} \\
}

\begin{document}

\maketitle

\begin{abstract}
  This paper presents a novel approach in autonomous robot control, named LLM-BRAIn, that makes possible robot behavior generation, based on operator's commands. LLM-BRAIn is a transformer-based Large Language Model (LLM) fine-tuned from Stanford Alpaca 7B model to generate robot behavior tree (BT) from the text description. We train the LLM-BRAIn on 8,5k instruction-following demonstrations, generated in the style of self-instruct using text-davinchi-003. The developed model accurately builds complex robot behavior while remaining small enough to be run on the robot's onboard microcomputer. The model gives structural and logical correct BTs and can successfully manage instructions that were not presented in training set. The experiment did not reveal any significant subjective differences between BTs generated by LLM-BRAIn and those created by humans (on average, participants were able to correctly distinguish between LLM-BRAIn generated BTs and human-created BTs in only 4.53 out of 10 cases, indicating that their performance was close to random chance). The proposed approach potentially can be applied to mobile robotics, drone operation, robot manipulator systems and Industry 4.0.
\end{abstract}

\section{Introduction}

The recent release of high-performance large language models (LLM), such as the Stanford Alpaca model in Open Source, has opened up enormous possibilities for researchers in related technical fields. Primarily, this is due to the ability to fine-tune transformer-based LLM to perform instruction-following tasks. It is now evident that GPT-like models will soon be widely used in the field of human-robot interaction. In this context, it is important to highlight that the integration of this technology is particularly relevant to the field of robotics, as it has a significant impact on the development and implementation of AI-driven robotic systems. Additionally, the issue of interaction between robots and humans is currently relevant.

Efforts to formalize robot commands and logic have been underway for quite some time. The priority of these efforts is to provide non-technical specialists with the ability to control robots in the workplace and interact with them in a convenient and logically understandable manner. The behavior tree (BT) approach is the advanced method for specifying robot logic. It allows complex robot behavior to be assembled from simple blocks (nodes). Moreover, the BT provides block interchangeability and maintains an unambiguous and understandable structure that can be stored in an XML file. Modern transformer-based LLM are well-suited for the task of creating strictly structured text files based on the specified logic and are now capable of even generating program code, which is a more complex task.

In this paper, we propose a new approach to autonomous robot control, which involves generating complex robot behavior in the BT format using a library of pre-written nodes based on human operator verbal descriptions. The generation of such behavior is achieved through a fine-tuning transformer-based LLM.

\section{Related Works}
\label{related}

In recent times, the transformer models have gained significant popularity, becoming increasingly prevalent in various domains since the concept was introduced in the groundbreaking paper "Attention is all you need" by \citet{vaswani2017attention}. These transformer models have shown remarkable performance in tasks such as language modeling, translation, and speech recognition, among others.

In particular, LLMs, which are based on the transformer architecture, have received tremendous attention in recent years. The rise in popularity has been facilitated by the emergence of models such as OpenAI's GPT2 and GPT3, whose features are discussed by \citet{gpt2} and \citet{gpt3} respectively. The introduction of ChatGPT in \citet{openai2022introducing} has significantly contributed to the growing popularity by bringing accessibility to GPT to a broader audience, taking the level of popularity to new heights. These models have demonstrated the ability to generate coherent and contextually appropriate text, making them well-suited for a wide range of natural language processing (NLP) tasks.

For a while, only those who collaborated with OpenAI had the opportunity to actively participate in the development of GPT models and witness their remarkable successes. However, the landscape changed with the advent of GPT-like LLM counterparts such as Google T5 described in paper \citet{t5} and Meta's LLaMa recently introduced in \citet{meta}, which offered alternative options that were not restricted to a single entity's control. These models have gained widespread usage across various NLP applications, encompassing language generation, question-answering, sentiment analysis, and more.

Numerous laboratories and industries with access to LLMs have been diligent in developing a methodology that uses GPT or its counterparts to develop methods for building robot behavior. One such work by \citet{driess2023palme} focuses on the development of PaLM-E, an embodied multimodal language model that makes a robot to follow instructions. Another work \citet{brohan2022rt1} presents a robotics transformer for real-world control at scale. Both of them are large models that requires significant computational resources and training data. Beyond that, Boston Dynamics has also been exploring the potential of GPT for building robot behavior. They have successfully trained a robot dog Spot to report on work results using GPT, as described in the article by \citet{petkauskas2023chatgpt}.

Furthermore, the recent availability of the Stanford Alpaca model presented in article \citet{alpaca} has become a turning point for LLM researchers worldwide. The repository of \citet{alpaca_lora_git} offers a comprehensive resource for individuals seeking to fine-tune Alpaca 7B model using the identical methodology employed by Stanford University. It encompasses all the necessary components and instructions to facilitate the fine-tuning process. Unlike other LLMs, the Stanford Alpaca model not only became accessible to researchers worldwide LLM performing close to GPT3, but also possessed characteristics that enabled it to run even on average power personal computers. The model's architecture allows for fine-tuning to perform instruction-following tasks, which opens up enormous possibilities for researchers in related technical fields.

Another work \citet{cao2023robot} proposes the use of GPT for generating robot behavior. Actually, in this work, the researchers have not generated full BT with LLM. They use an OpenAI product to fill fixed-structure BTs with behavior nodes. 
Since openAI products are not designed to generate robot behavior directly, the researchers had to make a limitation by fixing the BT structure and using only sequence and action nodes. Although this approach limits the modular structure advantages of BTs, the result was still promising. While the presented method for constructing a BT is a working approach, fine-tuning the LLM specifically for the task at hand usually yields much better results. This approach can be expected to generate various BTs without such limitations.

However, the generation of a diverse and extensive dataset with various BTs for robots of different structures and applications is necessary for fine-tuning LLM to achieve generation of complex BTs. While the approach of using Reinforcement Learning of from Human Feedback (RLHF) for LLM, as discussed by \citet{stiennon2022learning}, is widely acknowledged, generating a substantial dataset would necessitate employing specialists who possess expertise in constructing BT. This, in turn, renders the task highly challenging. Instead, this study uses the approach of \citet{wang2022selfinstruct} that was used in research of \citet{alpaca} to create a dataset using a text-davinci-003 model in the style of self-instruct. The resulting model can outperform the model used to generate the dataset in specific tasks because it is fine-tuned specifically to solve them.

\section{System Overview}
\label{system-overview}

\begin{figure}
  \centering
  \includegraphics[width=14cm,height=3.5cm]{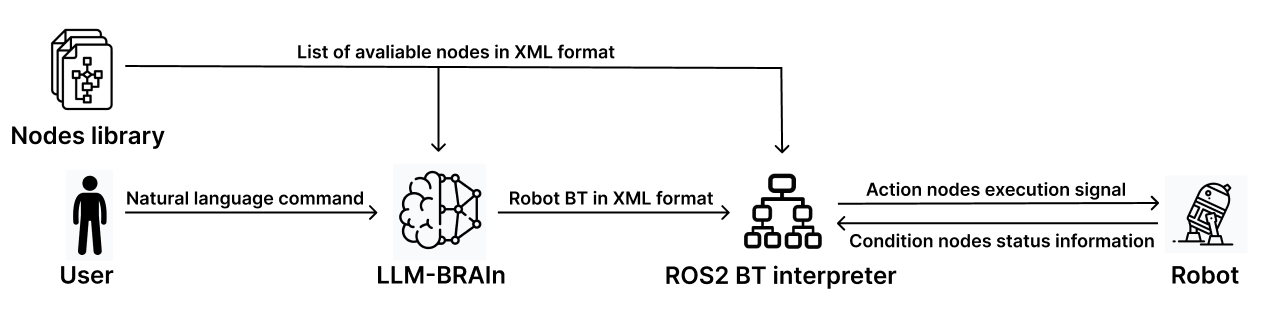}
  \caption{System architecture.}
  \label{system-architecture}
\end{figure}

The architecture of the system is presented in Fig.~\ref{system-architecture}. In the intended use case, the required hardware must include an onboard microcomputer capable of running the retrained LLM-BRAIn model with 7 billion weights. The hardware also includes all necessary sensors, actuators, and mechanical components of the robot required to perform the tasks described in its node library. The software system includes an application to run the LLM-BRAIn model, a BT interpreter that converts it into an executable file for the robot based on the Robot Operating System 2 (ROS2) which has become a standard in robotics as described by \citet{Macenski_2022}, and a node library that lists all the actions the robot can perform.

The operator can give the robot a command in natural language. Then, it is formulated as a request to the model and appended with the list of available nodes from the nodes' library. The resulting query is processed by the model. The output of the model, after converting special characters, is a generated BT in XML format. Previously, such BTs were written manually, but otherwise, this approach has become widespread in robotics because of its universality and versatility. Command execution based on the BehaviorTree.CPP library placed in the repository \citet{BTCpp} is done as an ROS2 node.

\section{Utilizing LLM Generated BTs to Define Robot Behavior}
\label{bt_for_robot}

\subsection{Advantages of Using Behavior Trees in Robotics}
The BT is a hierarchical structure used to represent robot tasks at an abstract level, offering an alternative to the state machine paradigm. BT as an approach to constructing robot behavior is discussed in \citet{Colledanchise_2018}. Formally, a BT is a directed rooted tree with leaf nodes responsible for task execution and branch nodes defining the control-flow logic. The leaf nodes are either Action or Condition nodes. The former specifies a primitive task and returns a Success signal when the task is completed. The latter is used to evaluate a Boolean condition, such as the satisfaction of a specific sensor reading. The most commonly used branch nodes are the Sequence and Fallback nodes. The Sequence node executes its child nodes sequentially until the first Failure signal is received. The Fallback node, on the other hand, executes its child nodes sequentially until the first Success signal is received. Using these four node types, a BT can achieve the same task execution as a state machine. But it has a lot of advantages. Modular structure allows to add, remove and replace nodes without necessary to reconstruct all structure.

BTs have become a widely used approach in robotics due to their intuitive and efficient control of robotic systems. They provide a structured way to represent and control the behavior of autonomous agents, making them well-suited for a wide range of robotic applications.

BT has been successfully applied in many robotics competitions and challenges, including the DARPA Robotics Challenge, RoboCup, and Eurobot. For example, in the DARPA Robotics Challenge, teams used BT to control their robots in tasks such as driving a car, opening doors, and using power tools, which was considered in the paper \citet{darpaBT}. In RoboCup, BT was used to control multi-robot systems in tasks such as soccer playing and search and rescue scenarios discussed by \citet{robocupBT}. In Eurobot, BT was used to control autonomous robots in tasks such as navigating a maze and manipulating objects considered in the paper of \citet{EurobotBT}.

\subsection{Advantages of Using Behavior Trees as Output of LLM}
Behavioral Trees, a hierarchical and modular structure, offer a promising solution for the development of transformer-based LLM. The structure enables the replacement of nodes with identical types. Ability to replace tokens make the structure well-suited for use as an output of the transformer. Additionally, the modularity and scalability of the BT structure allow for the easy addition of new nodes or modification of existing ones.

Another compelling feature of the BT structure is its option to use a subtree as part of a main tree. This means that generated BTs can be added to the node library and used as a part of more complex behavior. By operating in recursive mode, the model can build the BT first at the top level of abstraction and then descend to a lower level of abstraction, generating missing nodes from simpler components. This method allows for the creation of large and complex behavior structures while remaining within token length limitations in the model output. This approach positively impacts the model's performance and removes the limitations on the final size of the BT.

\section{Dataset Collecting}
\label{Dataset}

\subsection{Dataset Representation Format}
Let's consider how to represent a dataset for training an LLM model for BT generation. While the Stanford Alpaca model generates a sequence of words one by one and the dataset used for its training consists of complete text, for the instruction-following task, the text is divided into three parts using the "instruction:", "input:", and "output:" elements. The "input:" element is optional and is only used in some cases in the original Stanford work. We did not use this element in the task of generating BTs. Thus, each dataset sample consists of an instruction for building the robot's behavior and an output represented as a logically and structurally correct BT that allows the robot to execute the task. The instruction includes a general part that is common to all samples, "Write a behavior tree for the robot to execute the command using only available nodes." This part is necessary for further work when the LM will not only build BTs but also perform other tasks, such as correcting behavior, building subtrees, or requesting explanations or results of certain actions. After the unchangeable part of the instruction, there is a description of the required robot behavior. When generating this part of the samples using the text-davinci-003 model, special attention was paid to the naturalness of the request. It should be a simple and logically understandable command that a human could give, as it is the human who will give the command to the robot when using the fine-tuned model. The last part of the instruction includes a list of Action and Condition nodes that the robot can perform. This is a crucial element of the instruction because it is thanks to the restriction and specification of the list of available nodes that the generated BTs are executable on the target robot. In addition to Action and Condition nodes, the library sometimes contains SubTree nodes. These nodes work like Action nodes, but themselves are compiled BTs. Adding SubTree nodes allows generating the overall logic of the robot's behavior in several stages by generating missing elements from the available nodes. This approach avoids generating large constructions using LLM in one go, which would significantly increase the required memory, as the need to store attention between all sequence elements leads to a quadratic increase in the required memory with the increase in sequence length.

\subsection{Generating a Dataset Using Text-Davinci-003}
As previously noted, OpenAI's products are not intended for generating BTs based on descriptions. However, the text-davinci-003 model is capable of generating BTs of a robot's random behavior with different structures in the XML format. This capability serves as the starting point for generating a dataset. In the first request, we ask the model to generate a BT. Empirically, it has been found that this request needs additional clarifications that the BT must be executable by a mobile robot, have a peaceful application, and contain elements describing the robot's movement and manipulation of objects. This description can be modified for extrapolation to other types of robots. Additionally, certain structural and logical errors that the text-davinci-003 can make have been identified. By refining the requests, we were able to avoid these errors. In the second request, we ask text-davinci-003 to create a library of nodes that make up the generated BT, and we provide an example of the structure we want for the response. In the third request, we ask for a verbal description of the robot's behavior based on the generated BT, a task that text-davinci-003 performs well. Finally, we debug the three requests and use OpenAI's API to create a dataset by inserting the generated responses into the appropriate places in the samples. By following this approach, we created three datasets: for 1000 samples, 5000 samples, and 8500 samples. The quality criteria for such a dataset can be the diversity of presented samples within the target robot task, the correct structure of all BTs in the dataset, and the correspondence of the BT to the specified task.

\section{LLM-BRAIn 7B Model Fine-tuning}
\label{finetuning-brain}

\subsection{LLM-BRAIn Obtaining Approach}

\begin{figure}
  \centering
  \includegraphics[width=12cm,height=8cm]{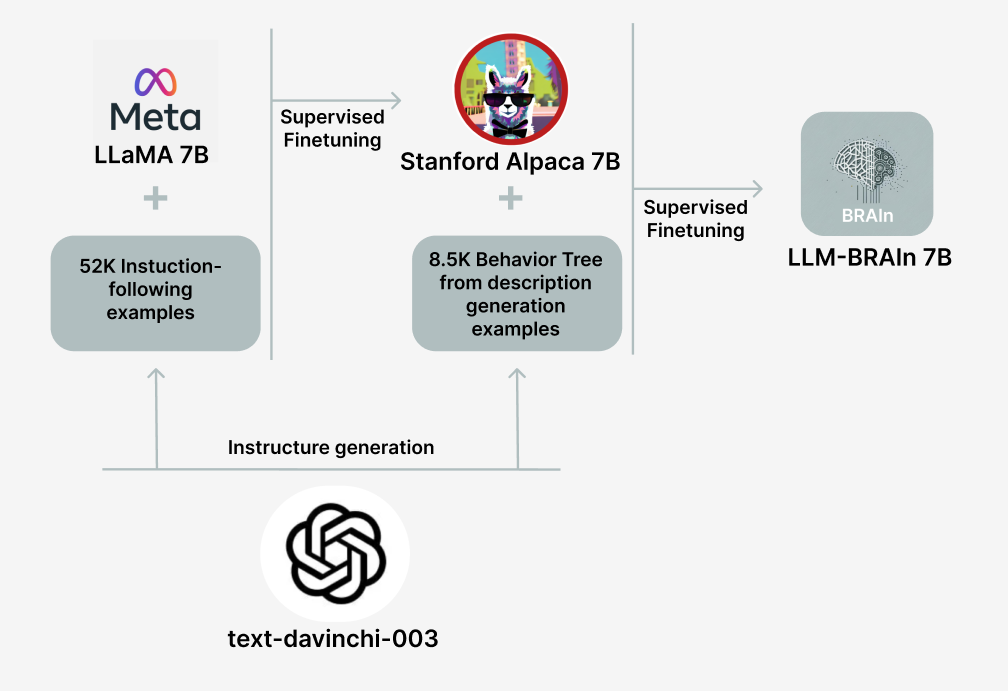}
  \caption{LLM-BRAIn fine-tuning overview.}
  \label{obtaining}
\end{figure}

The Stanford Alpaca 7B model is a language model that has been fine-tuned from the LLaMA 7B model using 52,000 instruction-following demonstrations. This fine-tuning approach has enabled the model to perform exceptionally well on instruction following tasks. However, constructing a BT for a robot is a challenging task for language models. To address this challenge, we further fine-tuned the model using examples of BT generation. Fig.~\ref{obtaining} illustrates how we obtained the LLM-BRAIn model.

Our goal in fine-tuning the model was twofold: first, to ensure that the fine-tuned model could generate a BT that satisfied all specified conditions, and second, to investigate the impact of dataset size on the quality of the generated BT. To achieve these goals, we employed the Parameter-Efficient Fine-Tuning approach. The PEFT methods allow you to efficiently and productively fine-tune the LLM without having to retrain the entire model, as discussed in \citet{PEFT}.

PEFT methods are particularly useful when tuning LLMs is too computationally expensive. Instead of tuning all the model's parameters, PEFT methods adjust only a small number of additional parameters, thus significantly reducing computational and storage costs. The PEFT method we employed to fine-tune the LLM-BRAIn model, is the Low-Rank Adaptation (LoRA) method, which adds low-rank matrices to transformer layers and adjusts only those matrices instead of the entire model. The application of the LoRA method to the LLM is described \citet{hu2021lora}

\subsection{Fine-tuning Process}

\begin{figure}
  \centering
  \includegraphics[width=11.5cm,height=4.5cm]{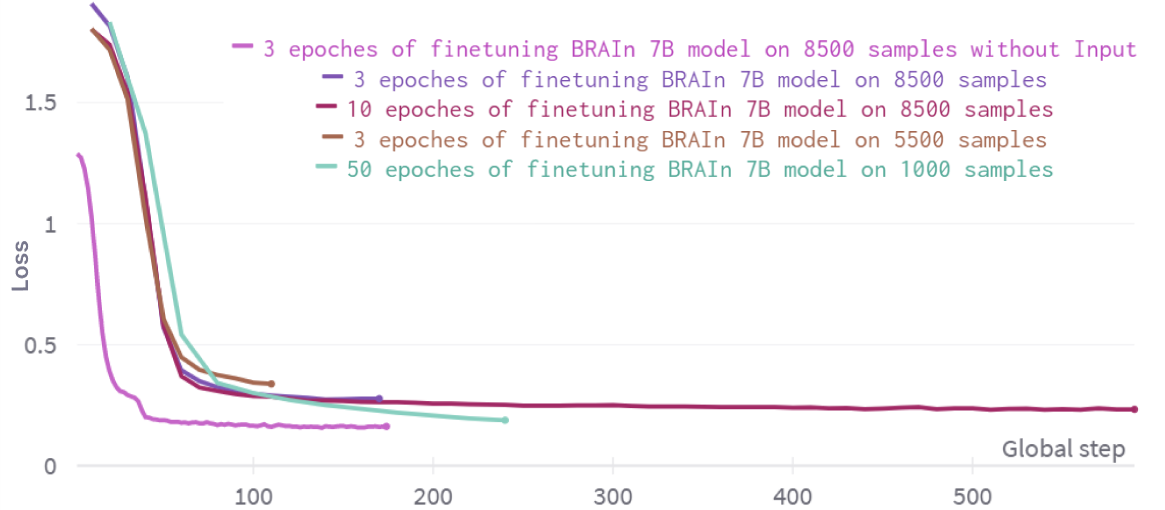}
  \caption{LLM-BRAIn fine-tuning curves.}
  \label{loss-fig}
\end{figure}

Fig.~\ref{loss-fig} shows the fine-tuning curves of the LLM-BRAIn model on different dataset sizes. It can be observed that the error graphs have a similar shape, allowing us to determine when to stop training as it no longer yields significant results. It was also noted that using only the "instruction:" part of the query in the dataset samples, without the use of "input:", positively impacted the training speed. A fine-tuning process was performed on one NVIDIA Tesla A100 graphics card with 80Gb of video memory. This amount of video memory allowed us to use batch size of 128 samples and micro batch size of 4 samples. However, our results could be repeated with less memory, if we change batch size to 32 samples and micro batch size to 1 sample. All basic hyperparameter values have been adopted as in the repository by \citet{alpaca_lora_git}. We assumed a learning rate of 3e-4, and the validation set size was 10\% of the entire dataset. The number of epochs required for training on the final dataset was 3. However, to achieve similar loss on different datasets, it was necessary to increase the number of training epochs for smaller datasets.

After fine-tuning the model, it was evaluated on random samples from the evaluation dataset. Following fine-tuning on 1000 samples, the model was able to reproduce the appearance of individual nodes in the required format, but was not able to construct complex BT logic. After increasing the dataset to 5500 samples and increasing its diversity, the model began to reproduce BTs that closely resembled those used in the industry, but still had logical errors in behavior, such as suggesting that the robot enter the door before opening it. Additionally, the model made non-trivial formatting errors, such as placing more than one node in the root of the BT. 

To address these issues, a dataset of 8500 samples was collected with a focus on tasks that previous models struggled with. As a result, the number of logical and formatting errors was minimized. Training the model on the final version of the dataset on one NVIDIA Tesla A100 graphics card with 80Gb of video memory was 8 hours. To obtain numerical values confirming the model's quality, experimental evaluations were performed.

\section{Experiment on Human Ability to Recognize Model-Generated Behavior Tree}
\label{experiment}

To assess the performance of the model, we conducted an evaluation of individuals' capacity to differentiate between LLM-BRAIn-generated BTs and manually authored ones by human experts. A similar investigation was conducted by \citet{gpt3} but with the distinction that in our study, all participants possessed knowledge regarding the underlying principles and construction of BTs.

\paragraph{Participants:}

We recruited a total of 15 participants, comprising both undergraduate and graduate students specializing in the Robotics track, to assess the quality of BT generation. Among the participants, ten individuals were members of local Eurobot team and regularly worked with BT. The remaining five participants had no prior exposure to BT and received explicit instructions on the underlying principles and construction techniques of BT. Prior to the commencement of the experiment, all participants provided their informed consent, thereby confirming their familiarity with the operational and structural aspects of robot BTs.

\paragraph{Procedure:}

\begin{figure}
  \centering
  \includegraphics[width=12cm,height=7.5cm]{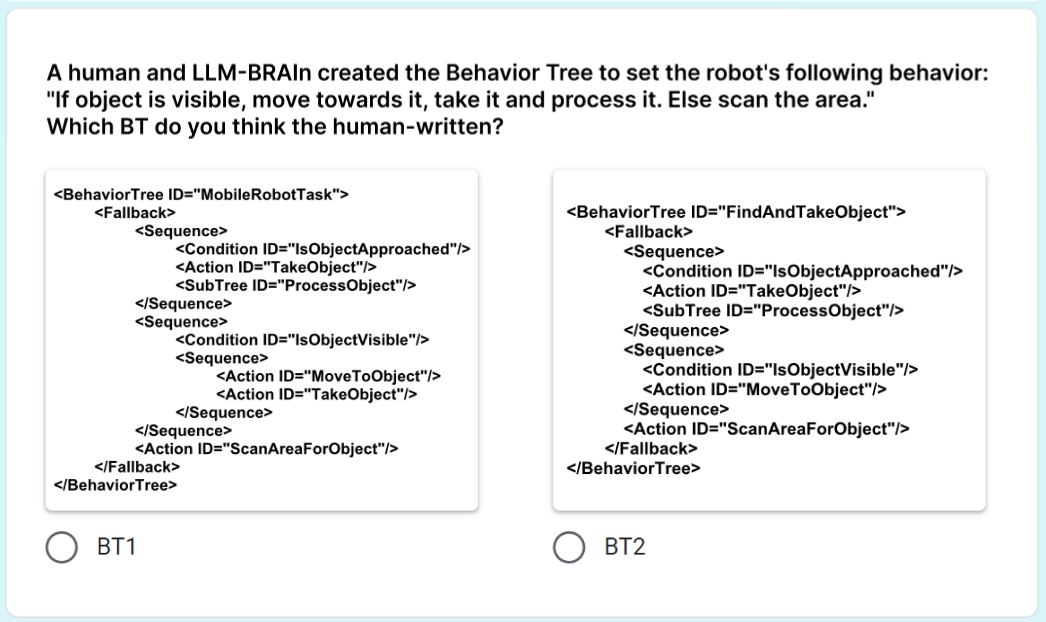}
  \caption{Example of task description and two solutions: human-written (BT1) and LLM generated (BT2) behavior trees.}
  \label{experiment-example}
\end{figure}

To conduct the experiment, a set of 10 BTs was generated using LLM-BRAIn. These BTs were designed to simulate various actions and interactions that a mobile robot could perform. Additionally, a separate set of 10 BTs was manually created by human experts, aiming to achieve similar functionalities and behaviors as the LLM-BRAIn generated BTs. Participants were presented with descriptions of the robot's behavior and pairs of BTs, consisting of one LLM-BRAIn generated BT and one human-created BT. The order of presentation was randomized to avoid any bias. Participants were then asked to evaluate the BTs and determine which one was created by the LLM-BRAIn and which one was human-created. They were instructed to rely on their subjective perception and any distinguishing features they could identify. An example question from the experiment is illustrated in Fig.~\ref{experiment-example}

\paragraph{Experimental Results:}

\begin{figure}
  \centering
  \includegraphics[width=12cm,height=5cm]{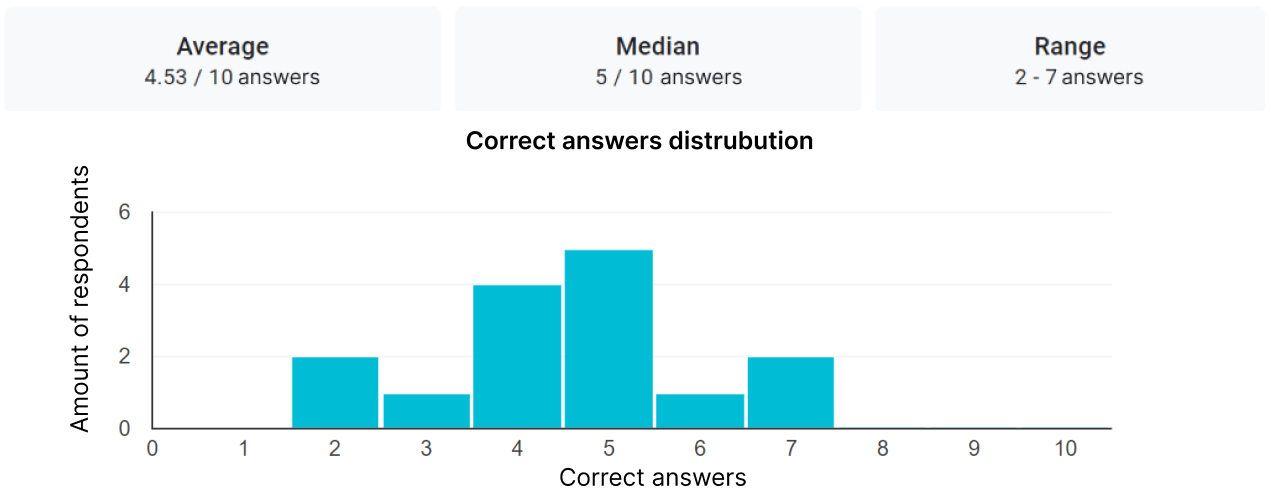}
  \caption{Correct answers given by the subjects.}
  \label{experiment-result}
\end{figure}

After the completion of the assessment phase for all pairs of Behavior Trees (BTs), the resulting data was collected and subjected to analysis. The findings of the experiment are visually depicted in Fig.~\ref{experiment-result}. The mean score of 4.53 correct answers out of a total of 10 suggests that participants' ability to differentiate between BTs generated by the LLM-BRAIn and those created by humans was comparable to random chance. 

In order to examine potential correlations between survey responses and specific questions, a one-way analysis of variance (ANOVA) was conducted with a significance level set at 5\%. The analysis revealed no statistically significant difference in users' perceptions of different questions (F = 0.75, p = 0.66 > 0.05). This suggests that there is no evidence to support the claim that the probability of a correct answer depends on the question. To determine whether there exist discernible distinctions between human-generated BTs and those generated by the LLM-BRAIn, a t-test was conducted with a significance level set at 5\% to evaluate the null hypothesis that they cannot be distinguished from each other. Under this null hypothesis, the average number of correct answers provided by the subjects would be 5 out of 10, and the average score would be 0.5. Based on our results (mean score = 0.453, t-statistic = -1.200, critical value = 2.145, p > 0.05), we have no reason to reject the null hypothesis, which suggests that the mean score is not significantly different from 0.5.

The user study, involving 15 volunteers, yielded no substantive indications of subjective disparities between LLM-BRAIn-generated BTs and human-generated BTs. The LLM-BRAIn model demonstrates the capability to generate robot behavior with result close to human-created BTs, at least in terms of subjective perception within the context of our experiment.

\section{Limitations}
\label{limitations}

When utilizing the LLM-BRAIn, it is imperative to acknowledge certain limitations that arise as a result of the specific functioning of the model. Firstly, the model's current training does not encompass the generation of subtrees. Consequently, the size of the generated BT is restricted by the capabilities of the device executing the LLM. In order to operate the model, it is necessary to store the attention weights between all tokens of the generated text in video memory or RAM. This constraint restricts the number of tokens available for generation and, ultimately, the size of the BT. However, this limitation will be mitigated in the future by incorporating the ability to recursively generate subtrees. 

Moreover, although the LLM-BRAIn is positioned as a comprehensive control system for robots, its application to specific devices may necessitate additional training of the model on BTs tailored to those particular devices, encompassing examples of their control logic. For instance, in order to control drones effectively, BTs containing flight logic examples should be presented in the training set. 

Another limitation of the system stems from the library of nodes. This library encompasses all feasible actions that the robot can execute, as well as all feasible conditions it can assess. To install the LLM-BRAIn on a given robot or system, this library must be individually composed. While this practice imposes constraints on the range of actions the robot can undertake, it also renders the behavior more predictable and consequently safer. If all nodes in the library are designed with safety as a primary concern, the behavior of the robot will likewise prioritize safety.  This approach minimizes the potential negative effects that a robot equipped with LLM-BRAIn can cause.

\section{Conclusion}
\label{conclusion}

In this paper, we present a novel approach for autonomous robot control, which enables the generation of robot behavior trees based on operator's commands. The LLM-BRAIn stands out due to its ability to accurately generate complex robot behavior while maintaining a compact size suitable for installation on an onboard robot microcomputer. The results of the user study did not reveal any subjective differences between the behavior trees generated by LLM-BRAIn and those generated by humans. Only in 45.3\% of cases, the subjects were able to distinguish between human-generated and LLM-BRAIn-generated behavior trees.

For future work, we have planned to enhance the model by enabling recursive generation of subtrees to expand the node library. Additionally, we aim to include functionality for the operator to change the behavior of the robot according to his requirements and request comments on it. Furthermore, our intentions include integrating LLM-BRAIn as a control system for mobile robots, robot manipulator systems, and drone swarms.

{
\small
 \bibliographystyle{unsrtnat}
 \bibliography{lib}

\begin{thebibliography}{22}
\providecommand{\natexlab}[1]{#1}
\providecommand{\url}[1]{\texttt{#1}}
\expandafter\ifx\csname urlstyle\endcsname\relax
  \providecommand{\doi}[1]{doi: #1}\else
  \providecommand{\doi}{doi: \begingroup \urlstyle{rm}\Url}\fi

\bibitem[Vaswani et~al.(2017)Vaswani, Shazeer, Parmar, Uszkoreit, Jones, Gomez,
  Kaiser, and Polosukhin]{vaswani2017attention}
Ashish Vaswani, Noam Shazeer, Niki Parmar, Jakob Uszkoreit, Llion Jones,
  Aidan~N. Gomez, Lukasz Kaiser, and Illia Polosukhin.
\newblock Attention is all you need, 2017.

\bibitem[Radford et~al.(2019)Radford, Wu, Child, Luan, Amodei, and
  Sutskever]{gpt2}
Alec Radford, Jeff Wu, Rewon Child, David Luan, Dario Amodei, and Ilya
  Sutskever.
\newblock Language models are unsupervised multitask learners.
\newblock 2019.

\bibitem[Brown et~al.(2020)Brown, Mann, Ryder, Subbiah, Kaplan, Dhariwal,
  Neelakantan, Shyam, Sastry, Askell, Agarwal, Herbert-Voss, Krueger, Henighan,
  Child, Ramesh, Ziegler, Wu, Winter, Hesse, Chen, Sigler, Litwin, Gray, Chess,
  Clark, Berner, McCandlish, Radford, Sutskever, and Amodei]{gpt3}
Tom~B. Brown, Benjamin Mann, Nick Ryder, Melanie Subbiah, Jared Kaplan,
  Prafulla Dhariwal, Arvind Neelakantan, Pranav Shyam, Girish Sastry, Amanda
  Askell, Sandhini Agarwal, Ariel Herbert-Voss, Gretchen Krueger, Tom Henighan,
  Rewon Child, Aditya Ramesh, Daniel~M. Ziegler, Jeffrey Wu, Clemens Winter,
  Christopher Hesse, Mark Chen, Eric Sigler, Mateusz Litwin, Scott Gray,
  Benjamin Chess, Jack Clark, Christopher Berner, Sam McCandlish, Alec Radford,
  Ilya Sutskever, and Dario Amodei.
\newblock Language models are few-shot learners, 2020.

\bibitem[OpenAI(2022)]{openai2022introducing}
OpenAI.
\newblock Introducing chatgpt.
\newblock \emph{OpenAI Blog}, 2022.
\newblock URL \url{https://openai.com/blog/introducing-chatgpt/}.

\bibitem[Raffel et~al.(2020)Raffel, Shazeer, Roberts, Lee, Narang, Matena,
  Zhou, Li, and Liu]{t5}
Colin Raffel, Noam Shazeer, Adam Roberts, Katherine Lee, Sharan Narang, Michael
  Matena, Yanqi Zhou, Wei Li, and Peter~J. Liu.
\newblock Exploring the limits of transfer learning with a unified text-to-text
  transformer, 2020.

\bibitem[Touvron et~al.(2023)Touvron, Lavril, Izacard, Martinet, Lachaux,
  Lacroix, Rozière, Goyal, Hambro, Azhar, Rodriguez, Joulin, Grave, and
  Lample]{meta}
Hugo Touvron, Thibaut Lavril, Gautier Izacard, Xavier Martinet, Marie-Anne
  Lachaux, Timothée Lacroix, Baptiste Rozière, Naman Goyal, Eric Hambro,
  Faisal Azhar, Aurelien Rodriguez, Armand Joulin, Edouard Grave, and Guillaume
  Lample.
\newblock Llama: Open and efficient foundation language models, 2023.

\bibitem[Driess et~al.(2023)Driess, Xia, Sajjadi, Lynch, Chowdhery, Ichter,
  Wahid, Tompson, Vuong, Yu, Huang, Chebotar, Sermanet, Duckworth, Levine,
  Vanhoucke, Hausman, Toussaint, Greff, Zeng, Mordatch, and
  Florence]{driess2023palme}
Danny Driess, Fei Xia, Mehdi S.~M. Sajjadi, Corey Lynch, Aakanksha Chowdhery,
  Brian Ichter, Ayzaan Wahid, Jonathan Tompson, Quan Vuong, Tianhe Yu, Wenlong
  Huang, Yevgen Chebotar, Pierre Sermanet, Daniel Duckworth, Sergey Levine,
  Vincent Vanhoucke, Karol Hausman, Marc Toussaint, Klaus Greff, Andy Zeng,
  Igor Mordatch, and Pete Florence.
\newblock Palm-e: An embodied multimodal language model, 2023.

\bibitem[Brohan et~al.(2022)Brohan, Brown, Carbajal, Chebotar, Dabis, Finn,
  Gopalakrishnan, Hausman, Herzog, Hsu, Ibarz, Ichter, Irpan, Jackson,
  Jesmonth, Joshi, Julian, Kalashnikov, Kuang, Leal, Lee, Levine, Lu, Malla,
  Manjunath, Mordatch, Nachum, Parada, Peralta, Perez, Pertsch, Quiambao, Rao,
  Ryoo, Salazar, Sanketi, Sayed, Singh, Sontakke, Stone, Tan, Tran, Vanhoucke,
  Vega, Vuong, Xia, Xiao, Xu, Xu, Yu, and Zitkovich]{brohan2022rt1}
Anthony Brohan, Noah Brown, Justice Carbajal, Yevgen Chebotar, Joseph Dabis,
  Chelsea Finn, Keerthana Gopalakrishnan, Karol Hausman, Alex Herzog, Jasmine
  Hsu, Julian Ibarz, Brian Ichter, Alex Irpan, Tomas Jackson, Sally Jesmonth,
  Nikhil~J Joshi, Ryan Julian, Dmitry Kalashnikov, Yuheng Kuang, Isabel Leal,
  Kuang-Huei Lee, Sergey Levine, Yao Lu, Utsav Malla, Deeksha Manjunath, Igor
  Mordatch, Ofir Nachum, Carolina Parada, Jodilyn Peralta, Emily Perez, Karl
  Pertsch, Jornell Quiambao, Kanishka Rao, Michael Ryoo, Grecia Salazar, Pannag
  Sanketi, Kevin Sayed, Jaspiar Singh, Sumedh Sontakke, Austin Stone, Clayton
  Tan, Huong Tran, Vincent Vanhoucke, Steve Vega, Quan Vuong, Fei Xia, Ted
  Xiao, Peng Xu, Sichun Xu, Tianhe Yu, and Brianna Zitkovich.
\newblock Rt-1: Robotics transformer for real-world control at scale, 2022.

\bibitem[Petkauskas(2023)]{petkauskas2023chatgpt}
Vilius Petkauskas.
\newblock Chatgpt injected into boston dynamics’ spot.
\newblock \emph{Cybernews}, 2023.
\newblock URL
  \url{https://cybernews.com/tech/chatgpt-google-boston-dynamics-spot/}.

\bibitem[Taori et~al.(2023)Taori, Gulrajani, Zhang, Dubois, Li, Guestrin,
  Liang, and Hashimoto]{alpaca}
Rohan Taori, Ishaan Gulrajani, Tianyi Zhang, Yann Dubois, Xuechen Li, Carlos
  Guestrin, Percy Liang, and Tatsunori~B. Hashimoto.
\newblock Alpaca: A strong, replicable instruction-following model.
\newblock \emph{CRFM Stanford University}, 2023.
\newblock URL \url{https://crfm.stanford.edu/2023/03/13/alpaca.html}.

\bibitem[Wang(2023)]{alpaca_lora_git}
Eric~J. Wang.
\newblock {Alpaca-LoRA}, 2023.
\newblock URL \url{https://github.com/tloen/alpaca-lora}.

\bibitem[Cao and Lee(2023)]{cao2023robot}
Yue Cao and C.~S.~George Lee.
\newblock Robot behavior-tree-based task generation with large language models,
  2023.

\bibitem[Stiennon et~al.(2022)Stiennon, Ouyang, Wu, Ziegler, Lowe, Voss,
  Radford, Amodei, and Christiano]{stiennon2022learning}
Nisan Stiennon, Long Ouyang, Jeff Wu, Daniel~M. Ziegler, Ryan Lowe, Chelsea
  Voss, Alec Radford, Dario Amodei, and Paul Christiano.
\newblock Learning to summarize from human feedback, 2022.

\bibitem[Wang et~al.(2022)Wang, Kordi, Mishra, Liu, Smith, Khashabi, and
  Hajishirzi]{wang2022selfinstruct}
Yizhong Wang, Yeganeh Kordi, Swaroop Mishra, Alisa Liu, Noah~A. Smith, Daniel
  Khashabi, and Hannaneh Hajishirzi.
\newblock Self-instruct: Aligning language model with self generated
  instructions, 2022.

\bibitem[Macenski et~al.(2022)Macenski, Foote, Gerkey, Lalancette, and
  Woodall]{Macenski_2022}
Steven Macenski, Tully Foote, Brian Gerkey, Chris Lalancette, and William
  Woodall.
\newblock Robot operating system 2: Design, architecture, and uses in the wild.
\newblock \emph{Science Robotics}, 7\penalty0 (66), may 2022.
\newblock \doi{10.1126/scirobotics.abm6074}.
\newblock URL \url{https://doi.org/10.1126%2Fscirobotics.abm6074}.

\bibitem[Faconti and Colledanchise(2019)]{BTCpp}
Davide Faconti and Michele Colledanchise.
\newblock {BehaviorTree.CPP}, 2019.
\newblock URL \url{https://github.com/BehaviorTree/BehaviorTree.CPP}.

\bibitem[Colledanchise and Ögren(2018)]{Colledanchise_2018}
Michele Colledanchise and Petter Ögren.
\newblock \emph{Behavior Trees in Robotics and {AI}}.
\newblock {CRC} Press, jul 2018.
\newblock \doi{10.1201/9780429489105}.
\newblock URL \url{https://doi.org/10.1201%2F9780429489105}.

\bibitem[Colledanchise and Ögren(2014)]{darpaBT}
Michele Colledanchise and Petter Ögren.
\newblock How behavior trees modularize robustness and safety in hybrid
  systems.
\newblock In \emph{2014 IEEE/RSJ International Conference on Intelligent Robots
  and Systems}, pages 1482--1488, 2014.
\newblock \doi{10.1109/IROS.2014.6942752}.

\bibitem[Safronov et~al.(2020)Safronov, Colledanchise, and Natale]{robocupBT}
Evgenii Safronov, Michele Colledanchise, and Lorenzo Natale.
\newblock Task planning with belief behavior trees, 2020.

\bibitem[Granosik et~al.(2016)Granosik, Andrzejczak, Kujawinski, Bonecki,
  Chlebowicz, Krysztofiak, Mirecki, and Gawryszewski]{EurobotBT}
Grzegorz Granosik, Kacper Andrzejczak, Mateusz Kujawinski, Rafal Bonecki,
  Łukasz Chlebowicz, Blazej Krysztofiak, Konrad Mirecki, and Marek
  Gawryszewski.
\newblock Using robot operating system for autonomous control of robots in
  eurobot, erc and robotour competitions.
\newblock \emph{Acta Polytechnica CTU Proceedings}, 6:\penalty0 11, 11 2016.
\newblock \doi{10.14311/APP.2016.6.0011}.

\bibitem[Pu et~al.(2023)Pu, Jain, Yin, and Kaplan]{PEFT}
George Pu, Anirudh Jain, Jihan Yin, and Russell Kaplan.
\newblock Empirical analysis of the strengths and weaknesses of peft techniques
  for llms, 2023.

\bibitem[Hu et~al.(2021)Hu, Shen, Wallis, Allen-Zhu, Li, Wang, Wang, and
  Chen]{hu2021lora}
Edward~J. Hu, Yelong Shen, Phillip Wallis, Zeyuan Allen-Zhu, Yuanzhi Li, Shean
  Wang, Lu~Wang, and Weizhu Chen.
\newblock Lora: Low-rank adaptation of large language models, 2021.

\end{thebibliography}
}

%%%%%%%%%%%%%%%%%%%%%%%%%%%%%%%%%%%%%%%%%%%%%%%%%%%%%%%%%%%%

\end{document}